\relax
\documentclass[letterpaper]{article} 
\usepackage{aaai21}  
\usepackage{times}  
\usepackage{helvet} 
\usepackage{courier}  
\usepackage[hyphens]{url}  
\usepackage{graphicx} 
\usepackage{natbib}  
\usepackage{caption} 
\usepackage{amsfonts}       
\usepackage{booktabs}       
\usepackage{bm}
\usepackage{amsmath}
\usepackage{color} 
\usepackage{algorithm,algorithmicx,algpseudocode}

\usepackage{graphicx}
\usepackage{caption}
\usepackage{subcaption}
\graphicspath{{./images/}}

\usepackage[disable]{todonotes}
\newcommand{\acN}[1]{\todo[size=\scriptsize, color=blue!30]{[AC] #1}}

\usepackage[switch]{lineno}

\title{Scaling Hamiltonian Monte Carlo Inference for Bayesian Neural Networks\\ with Symmetric Splitting}
\author{Adam D. Cobb, Brian Jalaian\\}
\affiliations{US Army Research Laboratory, Adelphi}
\begin{document}

\maketitle

\begin{abstract}
Hamiltonian Monte Carlo (HMC) is a Markov chain Monte Carlo (MCMC) approach that exhibits favourable exploration properties in high-dimensional models such as neural networks. Unfortunately, HMC has limited use in large-data regimes and little work has explored suitable approaches that aim to preserve the entire Hamiltonian. In our work, we introduce a new symmetric integration scheme for split HMC that does not rely on stochastic gradients. We show that our new formulation is more efficient than previous approaches and is easy to implement with a single GPU. As a result, we are able to perform full HMC over common deep learning architectures using entire data sets.
In addition, when we compare with stochastic gradient MCMC, we show that our method achieves better performance in both accuracy and uncertainty quantification.
Our approach demonstrates HMC as a feasible option when considering inference schemes for large-scale machine learning problems. 
\end{abstract}


\section{Introduction}
To this day, Hamiltonian Monte Carlo remains the gold standard for inference in Bayesian Neural Networks (BNNs) \cite{neal1995bayesian}. 
However, the challenge of scaling HMC to applications involving large data sets limits its wide-scale use. Instead, approaches that utilise stochastic gradients are preferred due to their ability to better scale with data set size. The challenge for these stochastic gradient approaches is often finding a compromise between scalability and the modelling of uncertainty. However, if we cannot afford to compromise on uncertainty performance, then any feasible way of performing HMC would be extremely attractive. This would allow us to leverage the properties of HMC in modern deep learning architectures that are already starting to play a key part in safety-critical applications such as in medical diagnosis \cite{leibig2017leveraging}, self-driving vehicles \cite{filos2020can}, and disaster response \cite{lu2020see}. 

The two common approaches for performing Bayesian inference in large-scale models are stochastic variational inference 
(e.g. \citet{graves2011practical, blundell2015weight, gal2016dropout})
and Markov chain Monte Carlo (MCMC).
The latter MCMC approach only became practical for large data with the introduction of 
Stochastic Gradient Langevin Dynamics (SGLD) \cite{welling2011bayesian}. The appeal of MCMC (including the stochastic gradient variant) is that once the samples have converged to the target distribution, we can be confident that we are sampling from the distribution of interest and not from an approximate variational distribution. As a result there now exist multiple stochastic gradient MCMC schemes for inference in BNNs \cite{chen2014stochastic, ding2014bayesian, zhang2020csgmcmc}.
In comparison to traditional implementations of MCMC, stochastic gradient approaches avoid using both a full likelihood model as well as a Metropolis-Hastings step. Instead, they tend to use a decaying learning rate and an approximation of the full likelihood.
\acN{Look in that paper for tall data and make sure you mention this Gaussian approximation for stochastic gradients if applicable.}
In contrast, we look to the original formulation of HMC and augment the Hamiltonian such that we can perform HMC over entire data sets.

In this work, we introduce a novel symmetric splitting integration scheme for HMC that is more robust than previous approaches and is easy to implement as part of 
the \texttt{hamiltorch} Python
package.\footnote{See \url{https://github.com/AdamCobb/hamiltorch} for the code repository.} Our approach allows us to take advantage of the superior high-dimensional exploration of HMC, by letting chains with long trajectory lengths explore the parameter space of neural networks. We show how we are able to perform HMC without stochastic approximations, and achieve results that are more robust to large data sets. In addition to improving on previous proposed splitting formulations, we introduce a realistic application of vehicle classification from acoustic data and show our novel symmetric split HMC inference scheme is also able to outperform its stochastic counterparts. In particular, our extensive analysis of uncertainty quantification shows the value of our approach over the stochastic MCMC baselines.

Our paper is structured as follows. Section \ref{sec:related_work} describes the related work. Section \ref{sec:theory} covers previous theory on HMC and split HMC, enabling us to introduce novel symmetric split HMC in Section \ref{sec:sym_split}. In Section \ref{sec:split_compare}, we compare our new scheme to previous splitting approaches, where we show how our new method scales more efficiently to large data. In Section \ref{sec:vehicle}, we compare novel symmetric split HMC with stochastic gradient approaches, demonstrating its superiority in uncertainty quantification. We then discuss the implications of our results in Section~\ref{sec:disc} and conclude in Section~\ref{sec:conc}.


\section{Related Work}\label{sec:related_work}

Augmenting the Hamiltonian to increase the feasibility of implementing 
full HMC is a well-known approach, yet it has been relatively untouched by the machine learning community in recent years, with the majority of effort focusing on stochastic gradient approaches (e.g. \citet{chen2014stochastic, ding2014bayesian, zhang2020csgmcmc}).
However, if we go back to the original work of \citet{neal1995bayesian}, we see the introduction of splitting according to data subsets. Neal's motivation for splitting was to improve exploration by taking advantage of data sets that are redundant, such that one can achieve a bigger effective step size. This splitting approach, which we will refer to as \textit{randomised splitting} due to its formulation, was not symmetrical and in subsequent work, \citet{neal2011mcmc} wrote that ``some symmetrical variation ... might produce better results.'' The other appearance of split HMC in the literature comes from \citet{shahbaba2014split}, where the Hamiltonian was split into two parts, such that one part was solved for analytically. This splitting approach facilitated larger step sizes to be taken and improved the exploration of the sampler. \citet{shahbaba2014split} also introduced the idea of splitting the data into two subsets, one for data lying near the decision boundary (first inferred by a MAP approximation) and the other for data far away from the boundary. This data splitting approach relies on both the symmetry of the log likelihood in logistic regression and on the ability to quickly perform a MAP approximation. Since these works, we are not aware of any further advances in split HMC that make it feasible to implement a full Hamiltonian on a single GPU. In Section \ref{sec:split_compare}, we will show that our symmetrical version of split HMC does better than randomised splitting, as had been predicted by \citet{neal2011mcmc}.

A further challenge that arises in HMC is when the target distribution's geometry prevents easy mixing. One can look to work that aims to alleviate these issues such as the framework introduced by \citet{girolami2011riemann} of Riemannian manifold HMC (RMHMC) or a more scalable approach to Bayesian hierarchical models by using the variant referred to as semi-separable HMC \cite{zhang2014semi}. Another related line of work involves sampling in a transformed parameter space and then using the inverse transform to go back to the original space \cite{marzouk2016introduction}. Recent work by \citet{hoffman2019neutra} actually demonstrates that this transformation can result in an equivalence to RMHMC, where the authors utilise normalising flows as their invertible transformation.
Instead of improving exploration by alleviating the detrimental effects of bad geometry, a further way to improve HMC's performance is to focus on samplers that automatically tune their hyperparameters (\citealp{hoffman2014no}; \citealp{betancourt2013generalizing}; \citealp{wang2013adaptive}). These adaptations to HMC all follow a different direction to our work, since they do not directly address the challenge of scaling to large data with BNNs. However, these improvements are also complementary as they have the potential to be combined with our work in the future.

Stochastic gradient approaches to MCMC have become the main way to perform MCMC in BNNs since their introduction by \citet{welling2011bayesian}. While SGLD naturally arises by combining a Robbins-Monro-type algorithm \cite{robbins1951stochastic} with Langevin dynamics, the equivalent formulation for HMC (i.e. SGHMC) is more challenging and requires a limiting Gaussian assumption and the introduction of a friction term \cite{chen2014stochastic}. Although SGHMC has seen wide use in the machine learning community since its introduction (e.g \citet{springenberg2016bayesian, gustafsson2020evaluating}), there are various works that criticise the approach.
\citet{bardenet2014towards} demonstrate that relying on a Gaussian noise assumption can result in poor performance. \citet{betancourt2015fundamental} further criticises the use of stochastic gradients in HMC, reporting that the only way to reduce bias with data subsampling is to ``subsample twice in a symmetric composition,'' where \citet{betancourt2015fundamental} directly refers to \citet{neal1995bayesian} and \citet{shahbaba2014split}. However, this proposition did not come with an applicable solution, but instead it came with a call to devise new ways of avoiding stochastic approximations. Despite some of the potential limitations of stochastic gradient MCMC approaches, there are now multiple implementations that enable successful inference in BNNs. For example, a popular direction of research has been to propose approaches that aim to cover multiple modes in the posterior, which has been motivated by the success of using deep ensembles \cite{lakshminarayanan2017simple, ashukha2020pitfalls}. Therefore, we now see schemes employing cyclic learning rates \cite{zhang2020csgmcmc} and utilising thermostats \cite{ding2014bayesian, Leimkuhler2016} to attain improved performance. As a final note, these approaches avoid relying on the Metropolis-Hastings step and instead decrease their step sizes to zero to ensure that they converge to the target distribution. 

Despite the existence of a few examples that have employed data subsampling with full HMC, there is still no widely-used scheme that can compete with stochastic gradient MCMC on a large data set. Furthermore, some works in the field have criticised stochastic gradient approaches and hinted at symmetric splitting approaches as a possible way forward. In this paper, we will show that symmetric splitting does offer a scalable and robust approach for inference in BNNs.


\section{Split Hamiltonian Monte Carlo}\label{sec:theory}

In this section we first provide a brief overview of HMC. We then describe the work by \citet{neal2011mcmc} and \citet{shahbaba2014split} and conclude by introducing our new variation of split HMC in Section \ref{sec:sym_split}.

\subsection{Hamiltonian Monte Carlo}
HMC is a gradient-based MCMC sampler that employs Hamiltonian dynamics to traverse the parameter space of models. We can use HMC to overcome the challenge of performing inference in highly complex Bayesian models by materialising samples from the unnormalised log posterior via the proportionality,
$$ p(\bm{\omega} \vert \mathbf{Y}, \mathbf{X}) \propto p(\mathbf{Y} \vert \mathbf{X}, \bm{\omega})\,p(\bm{\omega}),$$
which is derived from Bayes' rule. The model is a function of the parameters, $\bm{\omega}\in \mathbb{R}^D$, and is defined by the likelihood $p(\mathbf{Y} \vert \mathbf{X}, \bm{\omega})$ and the prior $p(\bm{\omega})$, where $\{\mathbf{X,Y}\}$ are the input-output data pairs. The prior encodes assumptions over the model parameters before observing any data. To take advantage of Hamiltonian dynamics in our Bayesian model, we can augment our system by introducing a momentum variable $\mathbf{p} \in \mathbb{R}^D$, such that we now have a log joint distribution, $\log [p(\bm{\omega},\mathbf{p})] = \log [p(\bm{\omega} \vert \mathbf{Y}, \mathbf{X})\,p(\mathbf{p})]$, that is proportional to the Hamiltonian, $H(\bm{\omega},\mathbf{p})$. If we let $p(\mathbf{p}) = \mathcal{N}(\mathbf{p} \vert \mathbf{0}, \mathbf{M})$, where the covariance $\mathbf{M}$ denotes the mass matrix,  our Hamiltonian can then be written as:\footnote{We are ignoring the constants.} 
\begin{equation}\label{eq:ham}
H(\bm{\omega},\mathbf{p}) = \underbrace{-\log [p(\mathbf{Y} \vert \mathbf{X}, \bm{\omega})\,p(\bm{\omega})]}_{\substack{\text{Potential Energy} \\ U(\bm{\omega})}} + \underbrace{^1/_2\  \mathbf{p}^{\top}\mathbf{M}^{-1}\mathbf{p}.}_{\substack{\text{Quadratic Kinetic Energy}\\ K(\mathbf{p})}}
\end{equation}
This form consists of a quadratic kinetic energy term derived from the log probability distribution of a Gaussian and a potential energy term, which is our original Bayesian model. We can then use Hamiltonian dynamics to collect samples from our posterior distribution, which we know up to a normalising constant. These equations of motion, 
\begin{align}\label{eq:HMC_deriv}
\frac{\mathrm{d}\bm{\omega}}{\mathrm{d}t} &= \frac{\partial H}{\partial \mathbf{p}} = \mathbf{M}^{-1}\mathbf{p}; \notag \\  \frac{\mathrm{d}\bm{p}}{\mathrm{d}t} &= - \frac{\partial H}{\partial \bm{\omega}} = \nabla_{\bm{\omega}}\log [p(\mathbf{Y} \vert \mathbf{X}, \bm{\omega})\,p(\bm{\omega})],
\end{align}
determine how trajectories on the parameter space propagate. However, solving these equations in practice requires simulation via discrete steps. The Stormer--Verlet or leapfrog integrator is an integration scheme that ensures reversibility by being symmetric in its sequencing, as well as being symplectic (i.e. volume preserving as is required for Hamiltonian systems). Therefore we can introduce the leapfrog integrator by following the series of transformations:
\begin{align}\label{eq:leapfrog}
&\mathbf{p}_{t + \epsilon / 2}  = \mathbf{p}_{t} + \frac{\epsilon}{2}\frac{\mathrm{d}\bm{p}}{\mathrm{d}t}(\bm{\omega}_{t}), \quad
\bm{\omega}_{t + \epsilon}  = \bm{\omega}_{t} + \epsilon \frac{\mathrm{d}\bm{\omega}}{\mathrm{d}t}(\mathbf{p}_{t + \epsilon / 2}),\notag \\
&\mathbf{p}_{t + \epsilon}  = \mathbf{p}_{t + \epsilon / 2} + \frac{\epsilon}{2}\frac{\mathrm{d}\bm{p}}{\mathrm{d}t}(\bm{\omega}_{t + \epsilon}),
\end{align}
where $t$ is the leapfrog step iteration and $\epsilon$ is the step size. We can then use this scheme to simulate $L$ steps that closely approximate the dynamics of the Hamiltonian system. Furthermore, for ease of notation, we can rewrite these transformations as a series of function compositions:
\begin{equation}
\phi^U_\epsilon : (\bm{\omega}_t, \mathbf{p}_t) \rightarrow (\bm{\omega}_t, \mathbf{p}_{t + \epsilon}), \ \ \phi^K_\epsilon : (\bm{\omega}_t, \mathbf{p}_t) \rightarrow (\bm{\omega}_{t+ \epsilon}, \mathbf{p}_t ),
\end{equation}
such that the overall symmetric mapping of Equation \eqref{eq:leapfrog} can be denoted as
$\phi^U_{\epsilon/2} \circ \phi^K_\epsilon \circ \phi^U_{\epsilon/2}$ \cite{strang1968construction}.

Finally, HMC is performed by sampling $\mathbf{p}_t \sim p(\mathbf{p})$ and then using Hamiltonian dynamics, starting from $\{\mathbf{p}, \bm{\omega} \}_{t}$, to propose a new pair of parameters $\{\mathbf{p}, \bm{\omega} \}_{t+L}$. We then require a Metropolis-Hastings step to either accept or reject the proposed parameters to correct for any possible error due to approximating the dynamics with discrete steps. For further details of HMC, please refer to \citet{neal2011mcmc}.

\subsection{Split Hamiltonian Monte Carlo}

The splitting of a Hamiltonian into a sum of its constituent parts has been previously described by both \citet{leimkuhler2004simulating} and \citet{sexton1992hamiltonian}. Its appearance in HMC in the context of data subsets first came in \citet{neal1995bayesian}, who introduced the randomised splitting approach.\footnote{This is explicitly described by \citet[Sec~5.1]{neal2011mcmc}.} The general idea is to split the Hamiltonian into a sum of $Q$ terms such that
\begin{equation}
H(\bm{\omega},\mathbf{p}) = H_1(\bm{\omega},\mathbf{p}) + H_2(\bm{\omega},\mathbf{p}) + \dots + H_{Q}(\bm{\omega},\mathbf{p}).
\end{equation}
This splitting is especially suited to the scenario, where the log-likelihood can be written as a sum over the data (i.e. data is independent), which is almost always the assumption for BNNs. Therefore \citet{neal1995bayesian} introduced the following split into $M$ data subsets:
\begin{equation}
H(\bm{\omega},\mathbf{p}) = \sum^M_{m=1}\left[U_m(\bm{\omega})/2 + K(\mathbf{p})/M + U_m(\bm{\omega})/2\right],
\end{equation}
where $U_m(\bm{\omega}) = -\log(p(\bm{\omega}))/M - \ell_m(\bm{\omega})$ and $\ell_m(\bm{\omega}) = \log p(\mathbf{Y}_m\vert\mathbf{X}_m, \bm{\omega} )$ is the log-likelihood over the data subset $\{\mathbf{X}_m,\mathbf{Y}_m\}$. Although the original purpose of this splitting was not with the intention of scaling to large data sets, its formulation nicely fits this scenario.  

The order of the splitting is important because the sequence of mappings corresponding the Hamiltonian dynamics of each $H_i$ must be symmetrical if we are to ensure the overall transition is reversible i.e. $H_i = H_{Q-i+1}$. Unfortunately, the above splitting follows the sequence $H_{3m-2}(\bm{\omega},\mathbf{p}) = H_{3m}(\bm{\omega},\mathbf{p}) = U_m(\bm{\omega})/2$ and $H_{3m-1}(\bm{\omega},\mathbf{p}) =K(\mathbf{p})/M$, where $Q = 3M$, such that the flow follows
\begin{equation}\label{eq:randomised_splitting}
\phi^H_{\epsilon} = \phi^{U_1}_{\epsilon/2} \circ \phi^{K/M}_\epsilon \circ \phi^{U_1}_{\epsilon/2} \circ \dots \circ \phi^{U_M}_{\epsilon/2} \circ \phi^{K/M}_\epsilon \circ \phi^{U_M}_{\epsilon/2}.
\end{equation}
This splitting is no longer symmetrical and therefore requires an extra step whereby the ordering of the $M$ subsets for each iteration is randomised. This randomisation ensures that the reverse trajectory and the forward trajectory have the same probability. 

Other than randomised splitting, \citet{shahbaba2014split} introduced the ``nested leapfrog", which followed a symmetrical formulation. The purpose of their ``nested leapfrog'' was to enable parts of the Hamiltonian to be solved either analytically or more cheaply. For their data splitting approach, they rely on a MAP approximation that must be computed in advance. This is then followed by an analysis of which data lies along the decision boundary. Their dependence on the quality of the MAP approximation as well as prior analysis of the data makes their approach less feasible when looking to scale to large data with BNNs. However, we offer our own data splitting baseline, which we refer to as naive splitting that is simply a nested leapfrog. This is the simplest way of building an integration scheme that both mimics full HMC and is symmetrical, i.e.
\begin{equation}\label{eq:naive_splitting}
\phi^H_{\epsilon} = \phi^{U_1}_{\epsilon/2} \circ \phi^{U_2}_{\epsilon/2} \circ \dots \circ \phi^{K}_\epsilon \circ \dots \circ \phi^{U_2}_{\epsilon/2} \circ \phi^{U_1}_{\epsilon/2}.
\end{equation}
This splitting is equivalent to implementing the original leapfrog in \eqref{eq:leapfrog}, where we simply evaluate parts of the likelihood in chunks and then sum them.

We have now introduced two baselines that split the Hamiltonian according to data subsets. In the next section we will introduce our new symmetrical alternative that results in a better-behaved sampling scheme.


\section{Novel Symmetric Split Hamiltonian Monte Carlo}\label{sec:sym_split}

Instead of following previous splitting approaches, we offer a symmetrical alternative that we will show to produce improved behaviour. We split our Hamiltonian into the same $M$ data subsets as for randomised splitting, however we now change the ordering and rescale the kinetic energy term by a value depending on the number of splits. Our symmetrical splitting is structured such that $H_{2m-1}(\bm{\omega},\mathbf{p}) = H_{2(2M - m)}(\bm{\omega},\mathbf{p}) = U_m(\bm{\omega})/2$ and $H_{2j}(\bm{\omega},\mathbf{p}) = H_{2(2M - j)-1}(\bm{\omega},\mathbf{p})  =K(\mathbf{p})/D$, where $D = (M-1)\times 2$, $m = 1, \dots, M$, and $j = 1, \dots, M-1$. As an example the overall transformation for $M=2$ would be written as
\begin{equation}
\phi^H_{\epsilon} = \phi^{U_1}_{\epsilon/2} \circ \phi^{K/2}_\epsilon \circ\phi^{U_2}_{\epsilon/2} \circ \phi^{U_2}_{\epsilon/2} \circ \phi^{K/2}_\epsilon \circ \phi^{U_1}_{\epsilon/2},
\end{equation}
where $D=2$, and as a further example for $M=3$:
\begin{align}
\phi^H_{\epsilon} = \phi^{U_1}_{\epsilon/2}& \circ \phi^{K/4}_\epsilon \circ\phi^{U_2}_{\epsilon/2} \circ \phi^{K/4} \circ \phi^{U_3}_{\epsilon/2}\notag \\ & \circ\phi^{U_3}_{\epsilon/2} \circ \phi^{K/4}_\epsilon \circ\phi^{U_2}_{\epsilon/2} \circ \phi^{K/4} \circ \phi^{U_1}_{\epsilon/2},
\end{align}
where $D=4$. More generally, Algorithm \ref{alg:split_leap} describes the novel symmetric split leapfrog scheme.

\begin{algorithm}
  \caption{Novel Symmetric Split Leapfrog Scheme}
  \label{alg:split_leap}
  \hspace*{\algorithmicindent} \textbf{Inputs:} $ \mathbf{p}_0  $, $ \bm{\omega}_0, \epsilon, L, M$
  \begin{algorithmic}[1]
      \State $D = 2\times(M-1)$  \Comment{Set the scaling factor for the parameter update step.}
      \For{$l$ in $1, \dots, L$}
      \For{$m$ in $1, \dots, M$}
      \State $\mathbf{p} = \mathbf{p} + \frac{\epsilon}{2}\frac{\mathrm{d}\bm{p}}{\mathrm{d}t}(\bm{\omega})$
      \If{$m < M$}
	 \State $\bm{\omega} = \bm{\omega} + \frac{\epsilon}{D} \frac{\mathrm{d}\bm{\omega}}{\mathrm{d}t}(\mathbf{p})$      
	 \EndIf
      \EndFor
       \For{$m$ in $M, \dots, 1$} \Comment{Note the reversal of the loop indexing.}
      \State $\mathbf{p} = \mathbf{p} + \frac{\epsilon}{2}\frac{\mathrm{d}\bm{p}}{\mathrm{d}t}(\bm{\omega})$
      \If{$m > 1$}
	 \State $\bm{\omega} = \bm{\omega} + \frac{\epsilon}{D} \frac{\mathrm{d}\bm{\omega}}{\mathrm{d}t}(\mathbf{p})$      
	 \EndIf
      \EndFor
      \EndFor
  \end{algorithmic}
\end{algorithm}
Unlike randomised splitting, our integrator is symmetrical and leads to a discretisation that is now reversible such that setting $\mathbf{p} = -\mathbf{p}$ results in the original $\bm{\omega}$. This property of reversibility is convenient for ensuring the Markov chain converges to the target distribution \cite[Page~244]{robert2013monte}.
\acN{Include a few comments on Neal's original motivation and on Shababa's}
\acN{Include Full HMC  with -p in appendix}
\acN{Could also cite Robert and Casella p.206 as a reversible kernel makes it easy to show that Markov Chain converges to CLT.}


\section{Comparison to Other Splitting Approaches}\label{sec:split_compare}
We now demonstrate that our new approach is more efficient than both naive splitting and randomised splitting.

\begin{figure*}
    \centering
    \begin{subfigure}[b]{0.245\textwidth}
        \includegraphics[width=\textwidth]{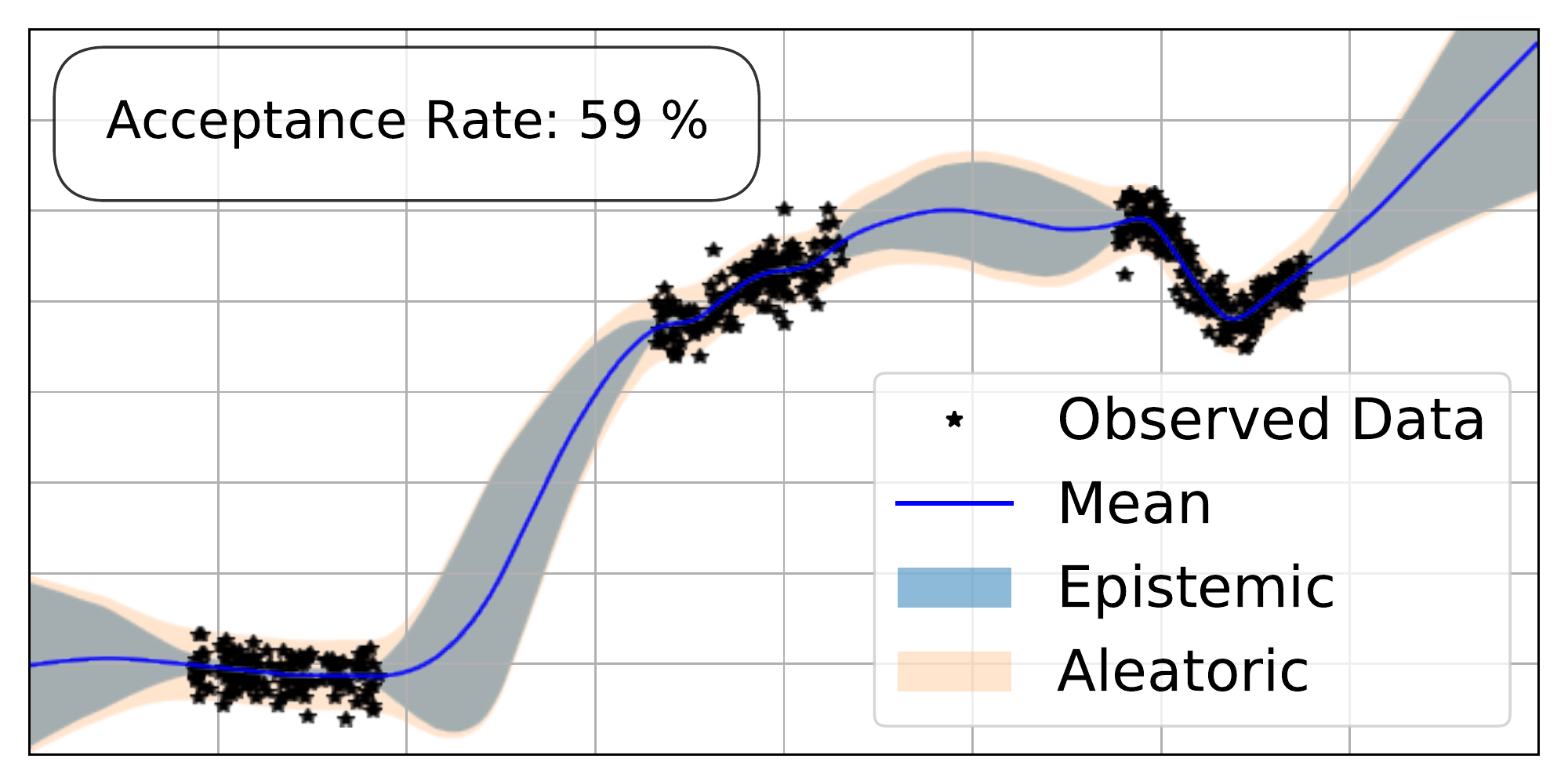}
        \caption{Full HMC}
        \label{fig:regression_full}
    \end{subfigure}
    \begin{subfigure}[b]{0.245\textwidth}
        \includegraphics[width=\textwidth]{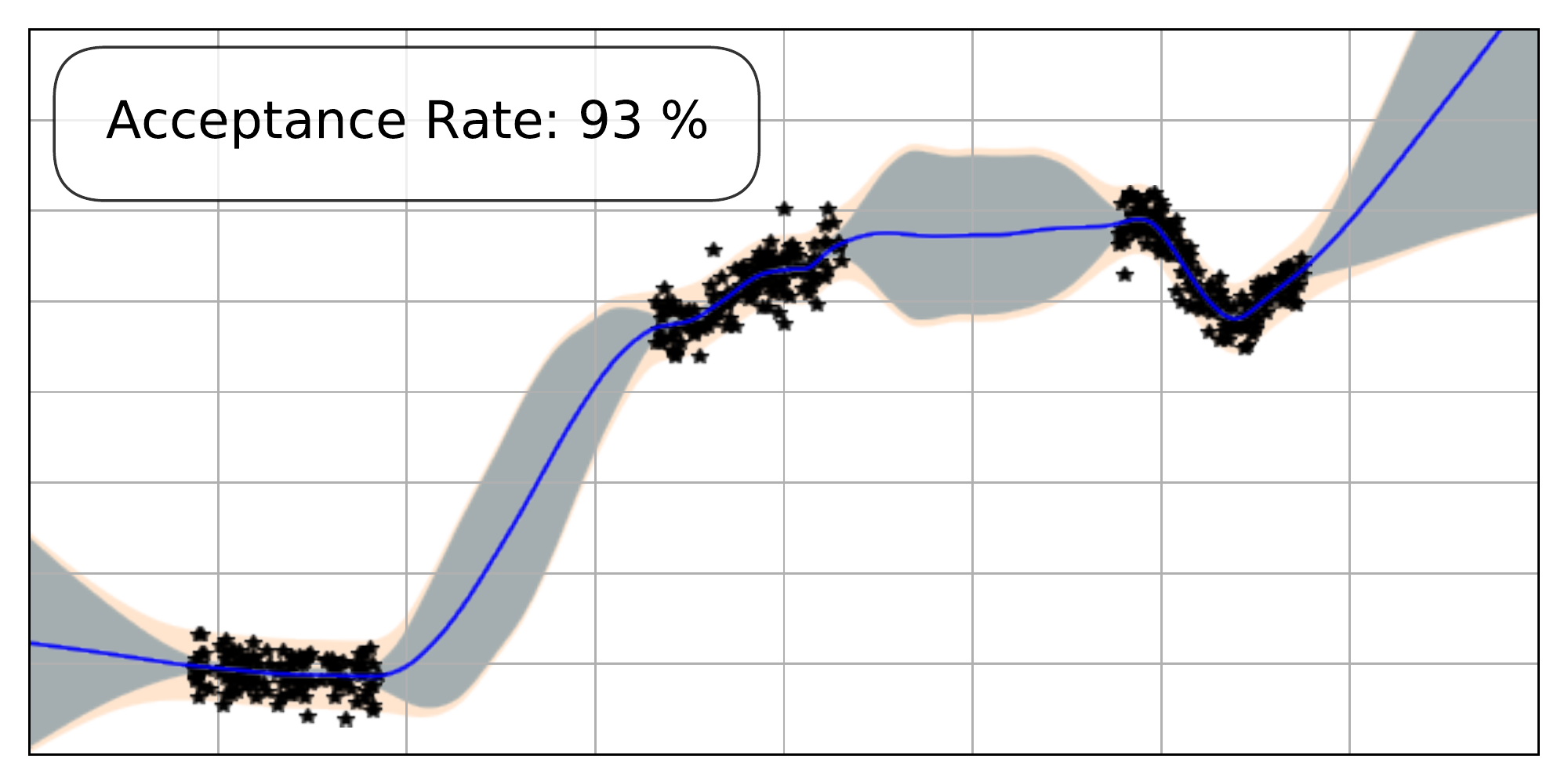}
        \caption{Novel symmetric split HMC}
        \label{fig:regression_novel_split}
    \end{subfigure}
    \begin{subfigure}[b]{0.245\textwidth}
        \includegraphics[width=\textwidth]{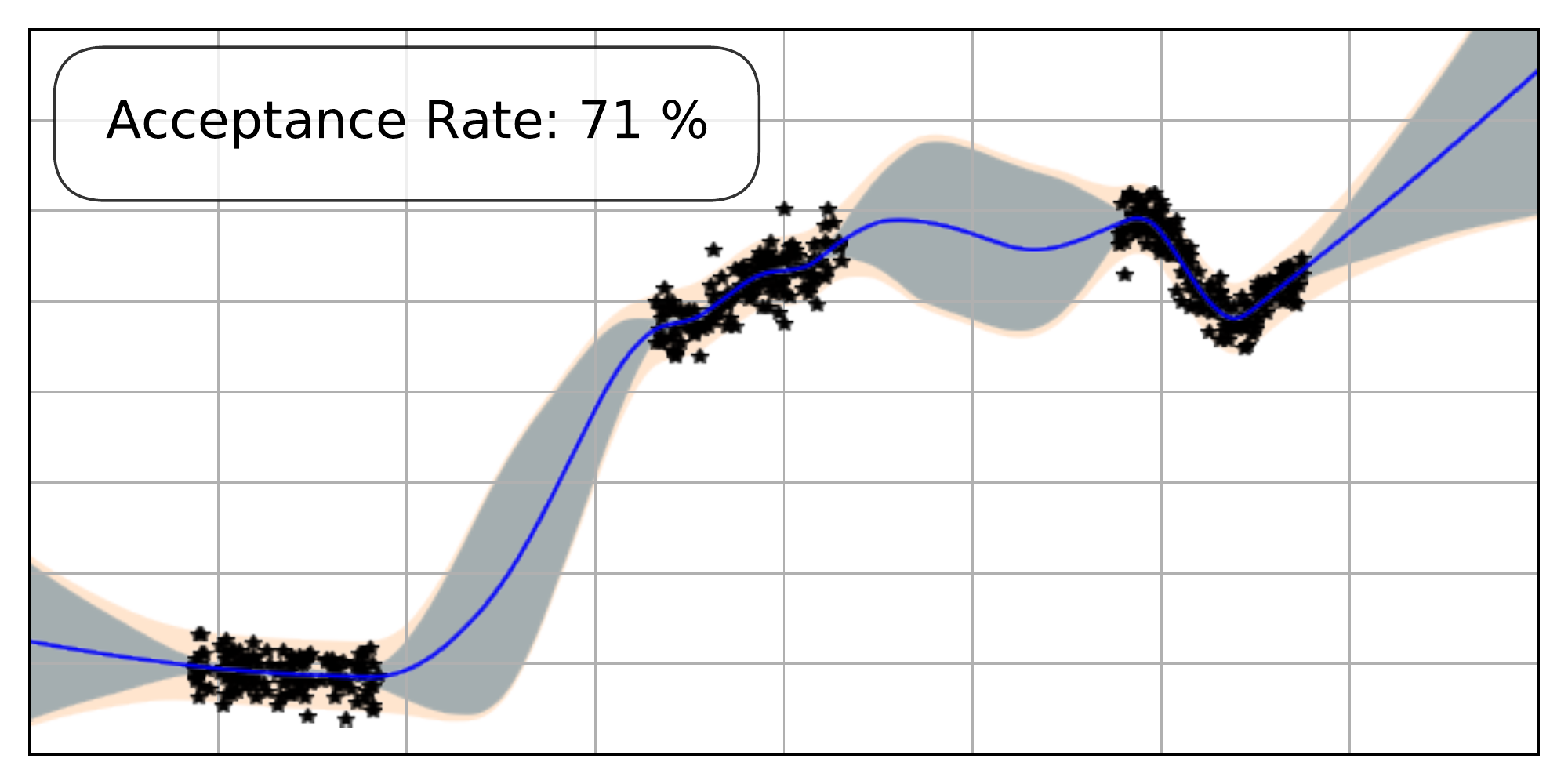}
        \caption{Randomised split HMC}
        \label{fig:regression_random}
    \end{subfigure}
    \begin{subfigure}[b]{0.245\textwidth}
        \includegraphics[width=\textwidth]{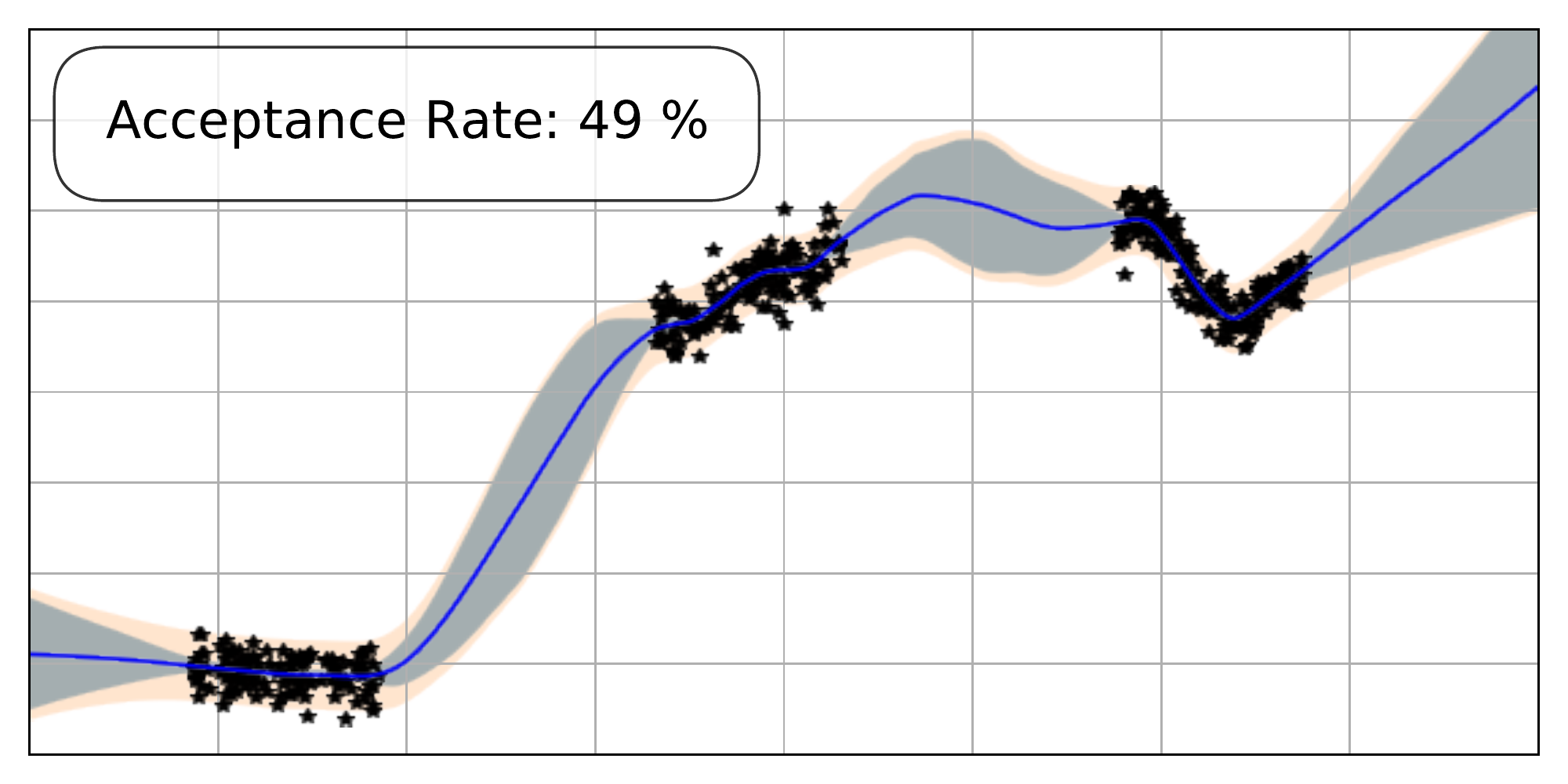}
        \caption{Naive split HMC}
        \label{fig:regression_naive}
    \end{subfigure}
    \caption{Regression example demonstrating the efficiency of novel symmetric HMC. A higher acceptance rate leads to better exploration and an increased epistemic uncertainty outside the range of the data. A lower acceptance rate corresponds to higher correlation between the samples. This higher correlation leads to a less efficient sampler for the same hyperparameter settings. E.g. note the narrower $2 \sigma$ epistemic credible intervals for both full and naive split HMC.}\label{fig:regression}
\end{figure*}

\subsection{Regression Example}
We illustrate regression performance across all approaches, where we use the simple 1D data set from \citet{Izmailov2019SubspaceIF} and set the architecture to a fully connected NN with $3$ hidden layers of $100$ units. Our model uses a Gaussian likelihood $p(\mathbf{Y}\vert \mathbf{X}, \bm{\omega}) = \mathcal{N}(\mathbf{f}(\mathbf{X};\bm{\omega}), \tau^{-1}\mathbf{I})$, where the output precision, $\tau$, must be tuned to characterise the inherent noise (aleatoric uncertainty) in the data. We implement a Gaussian process model with a Mat\'{e}rn $^3/_2$ kernel to learn this output precision with GPyTorch \cite{gardner2018gpytorch}.\footnote{In practice, $\tau$ can be tuned using cross validation as is the case for higher-dimensional problems.} For the splitting approaches we section the data into four subsets of $100$ training points each. All other hyperparameters are kept constant across the approaches to enable a fair comparison ($L=30$, $\epsilon = 5e^{-4}$, $\mathbf{M}=\mathbf{I}$, and $p(\bm{\omega}) = \mathcal{N}(\mathbf{0}, \mathbf{I})$).\footnote{These hyperparameters achieve a well-calibrated performance. For full HMC $30.5\ \%$ of the data lies outside the $1\sigma$ credible interval and $3.5\ \%$ for the $2\sigma$ interval.} Figure \ref{fig:regression} compares all four inference schemes. We include two standard deviation ($2\sigma$) credible intervals for both the aleatoric (including output precision) and epistemic uncertainty.

All inference schemes achieve comparable test log-likelihood scores (squared errors) and plateau after $200/1000$ samples are collected. However the acceptance rates across the schemes vary considerably, which can be seen from the results of Table \ref{tab:reg}. These results are calculated for ten randomly initialised HMC chains and show the mean and standard deviation for the acceptance rate, as well as the mean effective sample size (ESS).
Our novel symmetric splitting scheme achieves a significantly higher acceptance rate than all the other approaches. This higher acceptance rate increases mixing and results in an increased epistemic uncertainty outside the range of the data. 
This is shown by the wider epistemic credible intervals in Figure \ref{fig:regression_novel_split}. Conversely, a low acceptance rate leads to worse exploration and a higher correlation amongst the samples. The result is a less efficient sampler with narrower epistemic credible intervals in regions that do not contain data. We see this in Figure \ref{fig:regression_naive}, where the same hyperparameters lead to a collection of samples that expect less variation outside the range of the data. This overconfidence is undesirable and could possibly be overcome by reducing the step size or by increasing the total number of collected samples. However our approach of novel symmetric HMC shows that the current trajectory length ($L \times \epsilon$) achieves good results, and reducing this value for other approaches would increase computation for the same exploration.

\begin{table}[h!]
\small
  \caption{Regression example statistics calculated over $10$ HMC chains. The ESS was calculated using Pyro's in-built function \cite{bingham2019pyro}, followed by taking an average over the network's parameters ($\bm{\omega} \in \mathbb{R}^{10401}$). The acceptance rate is reported with its standard deviations. A higher mean ESS and a higher acceptance rate, demonstrate the better mixing performance from novel symmetric split HMC (as also seen in Figure \ref{fig:regression}). }
  \label{tab:reg}
  \centering
  \begin{tabular}{lcc}
    \toprule
    Inference Scheme & Acc. Rate & Mean ESS \\
    \midrule
    Full HMC & $0.63 \pm 0.06$ & $6.92$\\
    Naive Split HMC & $0.59 \pm 0.05$ & $6.77$   \\
    Randomised Split HMC & $0.73 \pm 0.07$ & $7.52$ \\
	Novel Sym. Split HMC &  $\mathbf{0.88\pm 0.04}$ & $\mathbf{7.72}$ \\
    \bottomrule
  \end{tabular}
\end{table}

\subsection{Classification Example}

We offer a further example to compare all four approaches where the difficulty of the task requires a larger model with two convolutional layers followed by two fully connected layers. This model has $38{,}390$ parameters. Our classification example uses the Fashion MNIST (FMNIST) data set \cite{Xiao2017FashionMNISTAN}, which we divide into a training set of $48{,}000$ images and a validation set of $12{,}000$ images. For the split HMC approaches, the training set is further split into three subsets of $16{,}000$. As for the regression example, all hyperparameters are set to the same values ($L=30$, $\epsilon = 2e^{-5}$, $\mathbf{M}=0.01\mathbf{I}$, and $p(\bm{\omega}) = \mathcal{N}(\mathbf{0}, \mathbf{I})$).

The results of this experiment can be seen in Table \ref{tab:fmnist_comp}, where novel symmetric split HMC achieves both a higher acceptance rate and higher mean effective sample size. This result is consistent with the previous regression example and further highlights the efficiency of our new splitting approach.

In this example, we see the advantage of using a splitting approach for tackling larger data tasks. For our specific hardware configuration (CPU: Intel i7-9750H; GPU: GeForce RTX 2080 with Max-Q), the maximum GPU memory usage with full HMC (using $48{,}000$ training images) is $7{,}928$ MB out of the available $7{,}982$ MB.
As a result, by splitting the data into three subsets, it would be possible to extend the current training set to $144{,}000$ training images without requiring a change in hardware. Therefore splitting makes it possible to perform HMC over much larger data sets, without the need for relying on stochastic subsampling.

\begin{table}
\small
  \caption{Classification example statistics calculated over $10$ HMC chains. The ESS was calculated using Pyro's in-built function \cite{bingham2019pyro}, followed by taking an average over the network's parameters ($\bm{\omega} \in \mathbb{R}^{38390}$). The acceptance rate is reported with its standard deviations. A higher mean ESS and a higher acceptance rate, demonstrate the better mixing performance of novel symmetric split HMC. }
  \label{tab:fmnist_comp}
  \centering
  \begin{tabular}{lccc}
    \toprule
    Inference Scheme & Acc. Rate & Mean ESS & Accuracy \\
    \midrule
    Full & $0.76 \pm 0.06$ & $6.26$ & $89.8 \pm 0.2$\\
    Naive Split & $0.72\pm 0.11$ & $6.21$ & $89.8 \pm 0.2$\\
    Randomised Split & $0.66 \pm 0.06$ & $6.24$ & $89.8\pm 0.2$ \\
	Novel Sym. Split & $\mathbf{0.89 \pm 0.02}$ & $\mathbf{6.37}$& $90.0 \pm 0.2$ \\
    \bottomrule
  \end{tabular}
\end{table}

\subsection{An Illustrative Example with a Larger Number of Splits}
As a final illustrative example, we also compare all four approaches on a small subset of $1{,}000$ CIFAR10 training images \cite{krizhevsky2009learning}. We use $10$ subsets of $100$ training images and show that split HMC is suitable for smaller batches. We run each HMC chain for $1{,}000$ iterations and burn the first $200$ samples (see Appendix \ref{ap:cifar10} for hyperparameters).
The results in Table \ref{tab:cifar10} follow the same pattern as the previous experiments, with a higher acceptance rate and higher mean ESS for novel symmetric split HMC compared to the others. Overall the results of Tables \ref{tab:reg}, \ref{tab:fmnist_comp}, and \ref{tab:cifar10} highlight the performance benefits of using our splitting approach, compared to previous approaches. These performance benefits become especially important in scenarios where splitting is a requirement of the hardware.

\begin{table}[h!]
\small
  \caption{Illustrative classification example performed over $1{,}000$ CIFAR10 training images with $5$ HMC chains per inference scheme. Here, we used $10$ data subsets to demonstrate the efficacy of our approach even with a larger number of smaller splits. Here we see that novel symmetric split HMC (NS) is more efficient for the same hyperparameter settings, with a higher acceptance rate, a higher mean ESS, and a higher accuracy. }
  \label{tab:cifar10}
  \centering
  \begin{tabular}{lccc}
    \toprule
    Inference Scheme & Acc. Rate & Mean ESS & Accuracy\\
    \midrule
    Full & $0.74 \pm 0.02$ & $73.19$ & $43.2 \pm 0.6$ \\
    Naive Split & $0.74 \pm 0.02$ & $73.90$ & $43.2 \pm 0.6$  \\
    Randomised Split & $0.60 \pm 0.02$ & $60.37$ & $43.1 \pm 0.5$\\
	Novel Sym. Split &  $\mathbf{0.83\pm 0.01}$ & $\mathbf{83.81}$ & $43.4 \pm 0.6$\\
    \bottomrule
  \end{tabular}
\end{table}


\section{Scaling HMC to a Real-World Example: Vehicle Classification from Acoustic Sensors}\label{sec:vehicle}

We will now show that our novel symmetric splitting approach facilitates applications to real-world scenarios, where the size of the data prevents the use of classical HMC. In our real-world example, the objective of the task is to detect and classify vehicles from their acoustic microphone recordings. 

\subsection{The Data Set}
The data consists of 223 audio recordings from the Acoustic-seismic Classification Identification Data Set (ACIDS).  ACIDS was originally used by \citet{hurd2002target} for harmonic feature extraction of ground vehicles for acoustic classification, identification, direction of arrival estimation and beamforming, but in this work we focus on acoustic classification. There are nine classes of vehicles, where each vehicle is recorded via a triangular array of three microphones.\footnote{Audio was recorded at a sampling rate of $1025.641$ Hz.}

In order to take advantage of the data structure from the three microphone sources, we transform each full time-series recording into the frequency domain using a short time Fourier transform (STFT), using the Scikit-learn default settings of \texttt{scipy.signal.spectrogram} \cite{JMLR:v12:pedregosa11a}. We randomly shuffle the recordings into eight cross-validation splits, where one is kept for hyperparameter optimisation. Once the audio recordings are divided,
they are split into smaller ($\approx 10$ s) chunks. We then work with the log power spectral density and build our training data by concatenating corresponding time chunks from all three microphones together into one spectrogram (e.g. see Figure \ref{fig:train_x} in Appendix \ref{ap:data}). Finally, the data is normalised using the mean and standard deviation of the log amplitude across the entire training data for each cross-validation split.

\subsection{Baselines}
We compare novel symmetric split HMC with Stochastic Gradient Descent (SGD), SGLD and SGHMC. All inference scheme hyperparameters are optimised via Bayesian optimisation using BoTorch \cite{balandat2019botorch}, adapted from the URSABench tool \cite{vadera2020ursabench}.

We use a neural network model that consists of four convolutional layers with max-pooling, followed by a fully-connected last layer. Importantly, we use Scaled Exponential Linear Units (SELUs) as the activation function \cite{klambauer2017self}, which we find yields an improvement over commonly-used alternatives such as rectified linear units. This is also seen by \citet{heek2019bayesian} for their stochastic gradient MCMC approach.

\acN{Run BO over these baselines}
\acN{Maybe add MC dropout}

\subsection{Classification Results}

Table \ref{tab:audio} displays the results of the experiment. We compare the four inference approaches and report their accuracy, Negative Log-Likelihood (NLL), and Brier score \cite{brier1950verification}, the last of which can be used to measure calibration performance. In our experimental set-up, we randomly allocate the data into seven train-validation splits and provide mean and standard deviations in Table \ref{tab:audio}. 
The result is that novel symmetric split HMC achieves an overall better performance compared to the stochastic gradient approaches. This demonstrates that one can perform HMC without using a stochastic gradient approximation on a single GPU and still achieve better accuracy and calibration.\footnote{We note that for our hardware, it was not possible to run full HMC on our GPU unless we reduced the training data by 53~\%.}

\setlength{\tabcolsep}{4pt}
\begin{table}
\small
  \caption{Vehicle classification results from acoustic data. Our novel symmetric split inference scheme outperforms in accuracy, NLL, and Brier score. The standard deviations are over seven randomised train-test splits.}
  \label{tab:audio}
  \centering
  \begin{tabular}{lccc}
    \toprule
    Method  & Accuracy & NLL & Brier Score \\
    \midrule
    SGD & $80.3 \pm 3.1$ & $ 0.72 \pm 0.15$ & $0.297 \pm 0.052$ \\
    SGLD & $78.6 \pm 3.3$ & $0.69 \pm 0.10$ & $0.307 \pm 0.043$ \\
    SGHMC & $82.6 \pm 3.1$ & $0.59 \pm 0.11$ & $0.252 \pm 0.042$ \\
    	NSS HMC & $\mathbf{84.4 \pm 2.1}$ & $\mathbf{0.51 \pm 0.05}$ & $\mathbf{0.228 \pm 0.027}$ \\
    \bottomrule
  \end{tabular}
\end{table}

\subsection{Uncertainty Quantification}
In addition to reporting the results in Table \ref{tab:audio}, we analyse the uncertainty performance across all cross-validation splits.
We will focus on two ways to analyse the quality of these results. 
First, we will focus on the predictive entropy as the proxy for uncertainty because this is directly related to the softmax outputs and is therefore the most likely to be used in practice. The posterior predictive entropy for each test datum $\mathbf{x}^*$ is given by the entropy of the expectation over the predictive distribution with respect to the posterior, $\mathcal{H}[\mathbb{E}_{\bm{\omega}}[p(\mathbf{y}^*\vert\mathbf{x}^*, \bm{\omega})]]$, which we will refer to via $\tilde{\mathcal{H}}$. 

\begin{figure}
\centering
    \includegraphics[width=.9\columnwidth]{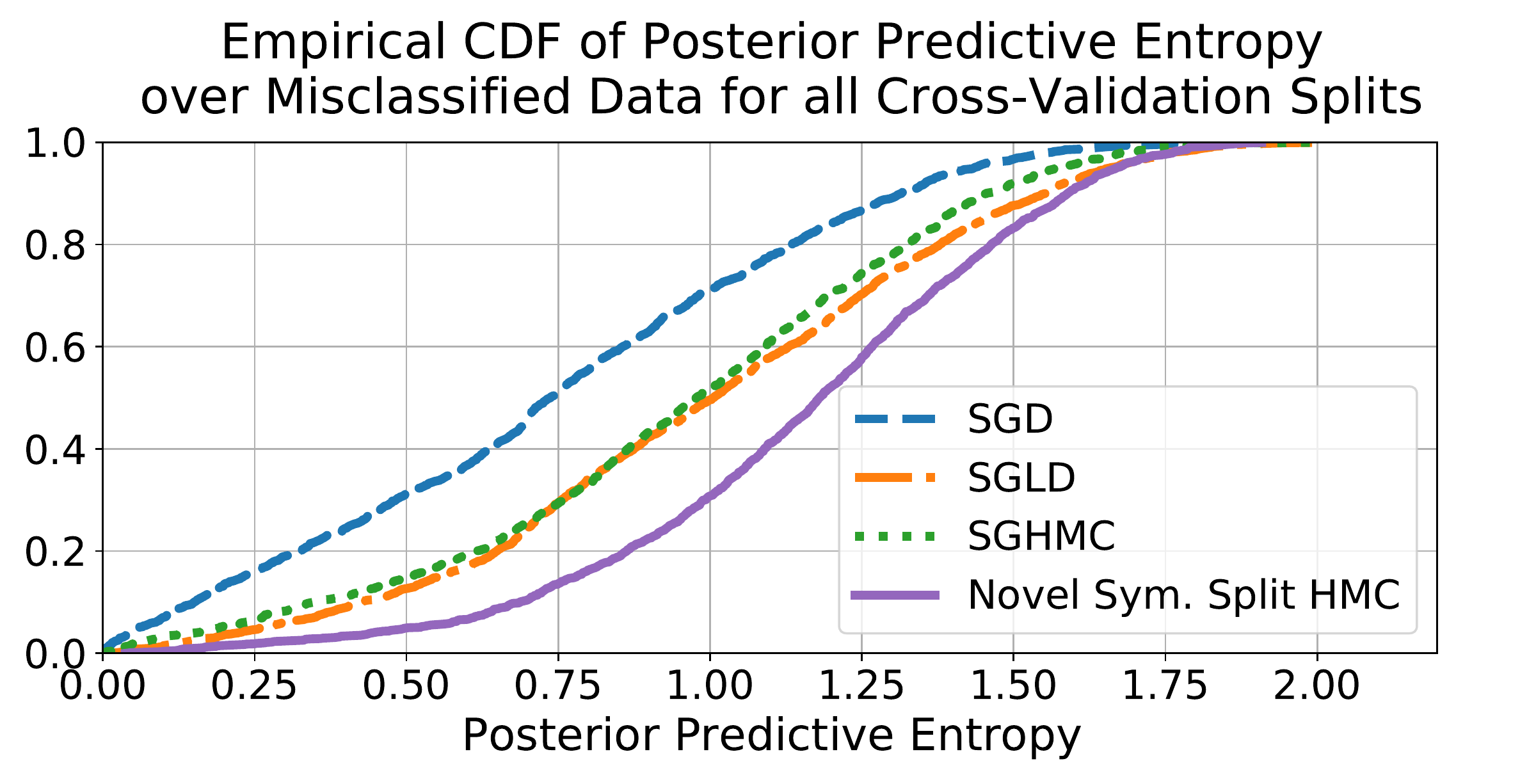}
    \caption{Cumulative posterior predictive entropy of misclassified data points. This plot shows that novel symmetric split HMC makes fewer high confidence errors than the other competing approaches. This is shown by the purple curve falling closer to the x-axis than the other curves.}
    \label{fig:cum_error}
\end{figure}

We can then plot the empirical cumulative distribution function (CDF) of all erroneous predictions across all cross-validation splits, as shown in Figure \ref{fig:cum_error}. It is desirable for a model to make predictions with high $\tilde{\mathcal{H}}$, when the predictions are wrong, which is the case for the misclassified data in Figure \ref{fig:cum_error}. Curves that follow this desirable behaviour remain close to the bottom right corner of the graph. Our new approach of novel symmetric split HMC behaves closer to the ideal behaviour in comparison to the baselines. This improved behaviour can be seen from the purple curve, which falls closer to the x-axis than the other curves. 

The second way that we will assess uncertainty is by relying on the mutual information between the predictions and model posterior. The mutual information can help distinguish between data uncertainty and model uncertainty, whereby our interest lies in the model uncertainty. Data points with high mutual information indicate that the model is uncertain due to the disagreement between the samples (this is in comparison to a model that is confident in its uncertainty, which would result in low mutual information). In the literature, the use of mutual information for uncertainty quantification can be seen in works by \citet{houlsby2011bayesian, gal2017deep} using Bayesian Active Learning by Disagreement (BALD) and via knowledge uncertainty in work by \citet{Depeweg2017DecompositionOU, malinin2020ensemble}.

\begin{figure}
    \centering
    \begin{subfigure}[b]{0.495\columnwidth}
        \includegraphics[width=\textwidth]{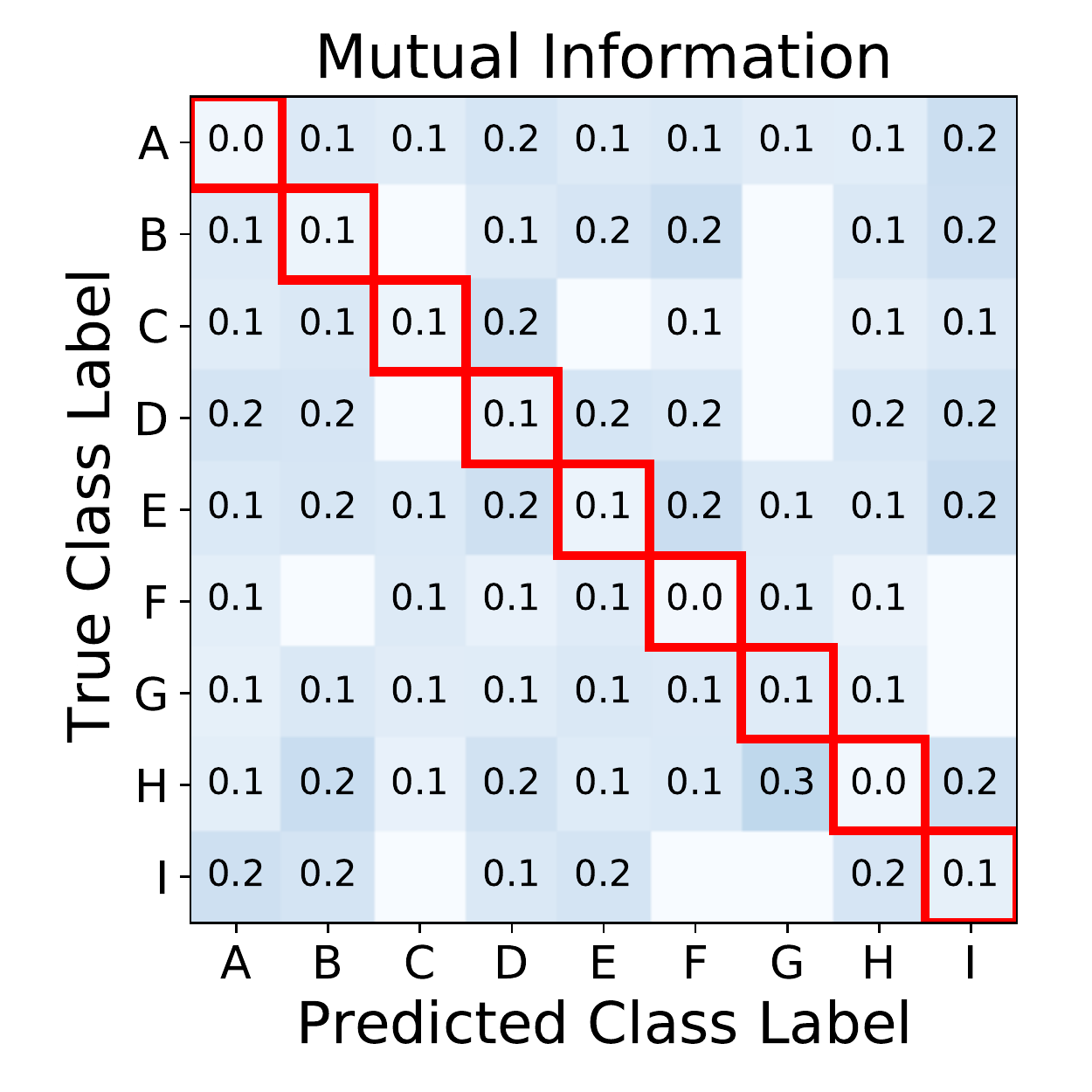}
        \caption{SGHMC}
        \label{fig:bald_sghmc}
    \end{subfigure}
    \hfill
    \begin{subfigure}[b]{0.495\columnwidth}
        \includegraphics[width=\textwidth]{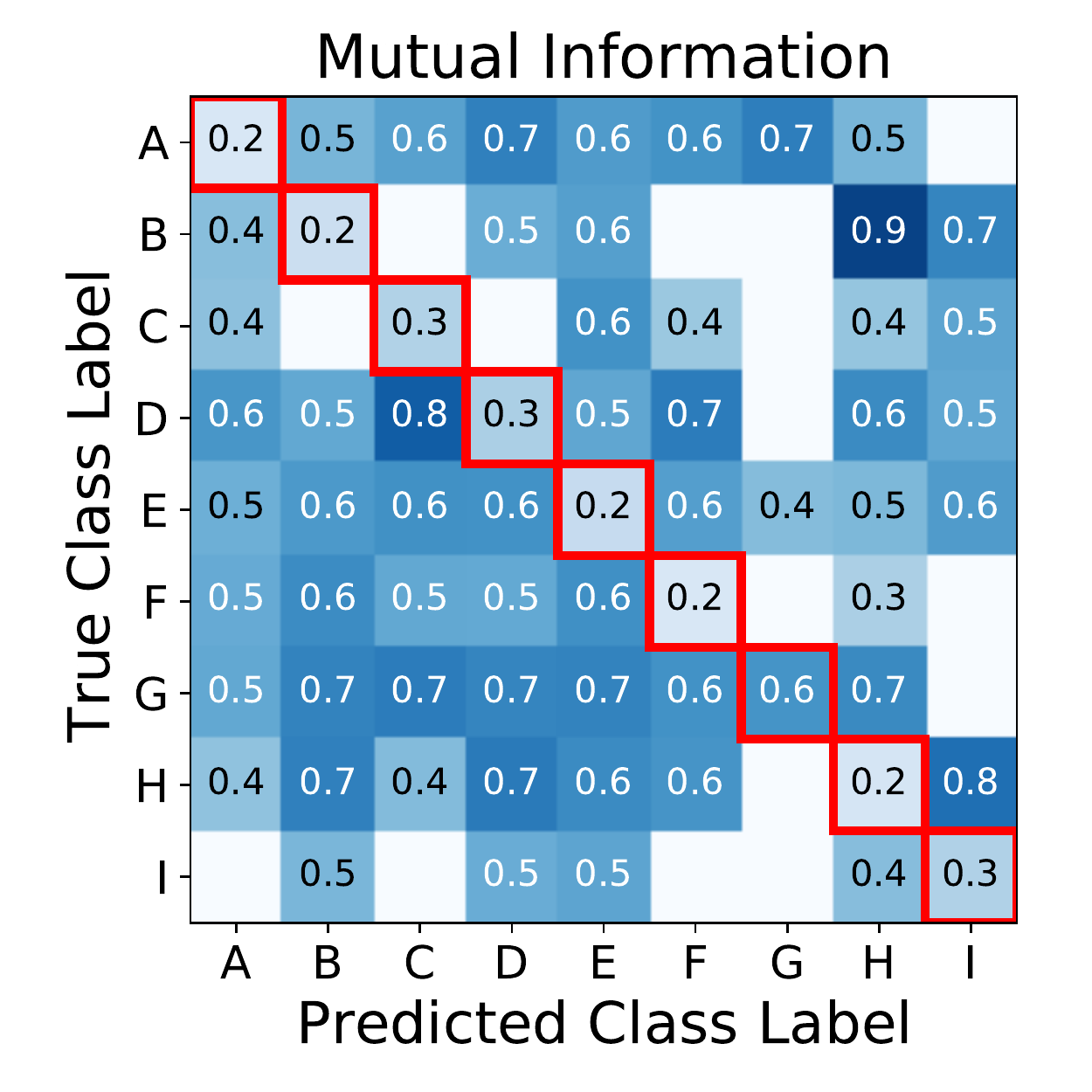}
        \caption{Novel sym. split HMC}
        \label{fig:bald_novel_split}
    \end{subfigure}
    \caption{``Confusion-style'' matrix showing average mutual information (MI) per category. Each square corresponds to the MI averaged over the number of test data corresponding to that box (boxes containing no data are blank). The diagonals (highlighted in red) indicate average MI over correct classifications, where low values are desirable. The off-diagonals indicate the average MI for erroneous predictions, where high values are desirable. (a) The matrix for SGHMC shows high MI everywhere, which is especially noticeable over the misclassifications. (b) Novel symmetric split HMC is more uncertain over its erroneous predictions and the difference between diagonals and off-diagonals is more obvious.}\label{fig:bald}
\end{figure}

To analyse mutual information, in Figure \ref{fig:bald}, we display ``confusion-style'' matrices for the top performing inference schemes according to Table \ref{tab:audio}, SGHMC and novel symmetric split HMC.
Each square in the matrix contains the average mutual information over all the data corresponding to that square across all the cross-validation splits. Low values along the diagonal are desirable because they correspond to confident predictions for correct classifications. However, low values on the off-diagonals are especially undesirable as they correspond to erroneous, highly confident predictions. When we compare SGHMC of Figure \ref{fig:bald_sghmc} to novel symmetric split HMC of Figure \ref{fig:bald_novel_split}, we see the advantages of our approach. The off-diagonals for SGHMC indicate that the model is making errors with little warning to the user that these errors actually exist. Furthermore, there is a lot of overlap between the average mutual information between the correct and wrong predictions. This overlap would make it hard to alert a user of any possible erroneous prediction. In comparison, our novel symmetric split approach shows little overlap between the off-diagonal values and the correct predictions on the diagonals. This near-separability can help to distinguish erroneous predictions by their high uncertainty.

\section{Discussion}\label{sec:disc}

There are many challenges associated with performing HMC over large hierarchical models such as BNNs. Our work makes strides in the right direction but there are further areas to explore. As alluded to in Section \ref{sec:related_work}, there are techniques that can be employed to improve hyperparameter optimisation. For example, in this paper we have assumed the mass matrix to be diagonal with one scaling factor, which may not be an optimal choice. Future work that utilises geometrically-inspired theory, such as metrics derived by \citet{girolami2011riemann} or \citet{hoffman2019neutra}, may further improve the current method. 
Another challenge with MCMC approaches is knowing when enough samples have been collected such that the samples provide a good representation of the target distribution. In high-dimensional models like neural networks, chains may take a long time to converge and it is important to build reliable metrics for convergence such as observing the effective sample size, plotting the log-posterior density of the samples, and plotting the cumulative accuracy (e.g. see Appendix \ref{ap:clas}). Automatically building these diagnostics into libraries can save computation time. 

\section{Conclusion}\label{sec:conc}
In this work we have shown the advantage of preserving the entire Hamiltonian for performing inference in Bayesian neural networks. In Section \ref{sec:split_compare} we provided two classification tasks and one regression task. We showed novel symmetric split HMC is better suited to inference in BNNs compared to previous splitting approaches. These previous approaches did not have the same efficiencies as our novel symmetric split integration scheme.
We then provided a real-world application in Section \ref{sec:vehicle}, where we compared novel symmetric split HMC with two stochastic gradient MCMC approaches. For this acoustic classification example, we were able to show that our new method outperformed stochastic gradient MCMC, both in classification accuracy and in uncertainty quantification. In particular, the analysis of the uncertainty quantification showed novel symmetric split HMC achieved a lower confidence for its misclassified labels, whilst also achieving a better overall accuracy. In conclusion, we have introduced a new splitting approach that is easy to implement on a single GPU. Our approach is better than previous splitting schemes and we have shown it is capable of outperforming stochastic gradient MCMC techniques.


\section*{Ethics Statement}

Uncertainty quantification in machine learning is vital for ensuring the safety of future systems.
We believe improvements to approximate Bayesian inference will allow future applications to operate
more robustly in uncertain environments. These improvements are necessary because the future will consist of a world with more automated systems in our surroundings, where these systems will often be operated by non-experts. 
Techniques like ours will ensure these systems are able to provide interpretable feedback via safer, better-calibrated outputs. Of course, further work is needed to ensure that the correct procedures are fully in place to incorporate techniques, such as our own, in larger pipelines to ensure the fairness and safety of overall systems.

\section*{Acknowledgements}
We would like to thank Tien Pham for making the data available and Ivan Kiskin for his great feedback. ACIDS (Acoustic-seismic Classification Identification Data Set) is an ideal
data set for developing and training acoustic classification and
identification algorithms. ACIDS along with other data sets can be obtained
online through the Automated Online Data Repository (AODR) \citep{bennett2018cloud}.
Research reported in this paper was sponsored in part by the CCDC Army Research Laboratory. The views and conclusions contained in this document are those of the authors and should not be interpreted as representing the official policies, either expressed or implied, of the Army Research Laboratory or the U.S. Government. The U.S. Government is authorized to reproduce and distribute reprints for Government purposes notwithstanding any copyright notation herein.


\bibliographystyle{aaai}
\bibliography{splitHMC}


\clearpage

\section{Appendix}
\appendix

\section{CIFAR10 Classification Experiment}\label{ap:cifar10}
\paragraph{Model Architecture: } The model starts with two convolutional layers, where each layer is followed by SELU activations and ($2\times 2$) max-pooling. Both layers have a kernel size of $5$, where the number of output channels of the first layer is $6$ and the second is $16$. The next three (and final) layers are fully connected and follow the structure $[400, 120, 84, 10]$, with SELU activations. 

\paragraph{Hyperparameters: } $L=20$; $\epsilon = 5e^{-6}$; $\mathbf{M}=1e^{-5}\mathbf{I}$; $p(\bm{\omega}) = \mathcal{N}(\mathbf{0}, \tau^{-1} \mathbf{I})$, (with $ \tau = 100$); number of splits $=10$ (each of batch size $100$); number of samples $= 1000$; burn $= 200$.

\section{Vehicle Classification from Acoustic Sensors}
\subsection{Data}\label{ap:data}

\begin{figure}[h]
\centering
    \includegraphics[width=0.4\textwidth]{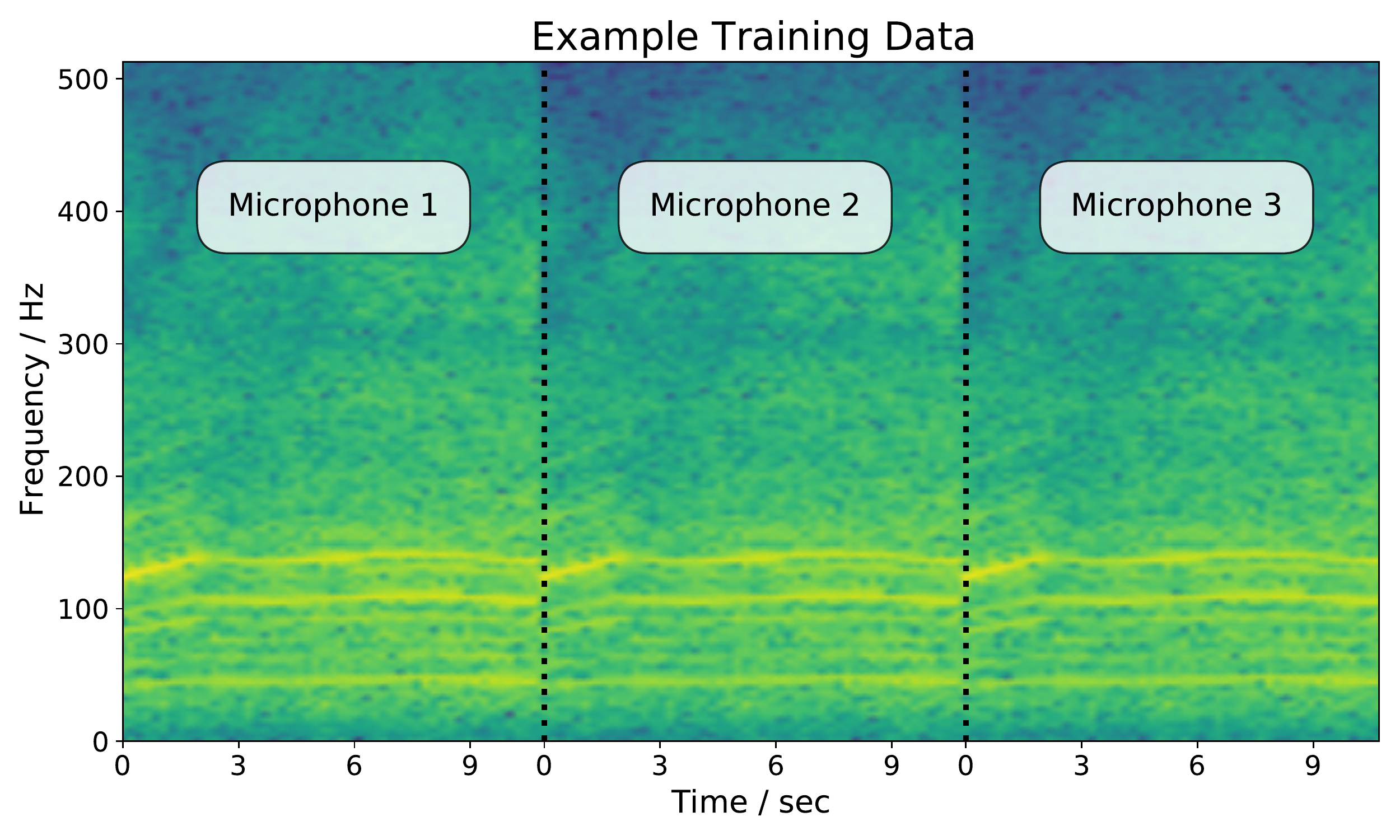}
    \caption{An example of a single input datum. The spectrograms from all three microphones (aligned in time) are concatenated into one image which is then passed into the CNN. The total $129 \times 150$ array has a resolution of $4.0$ Hz in the vertical axis and a resolution of $0.22$ seconds in the horizontal axis.}
    \label{fig:train_x}
\end{figure}

\begin{figure}[h]
\centering
    \includegraphics[width=0.98\columnwidth]{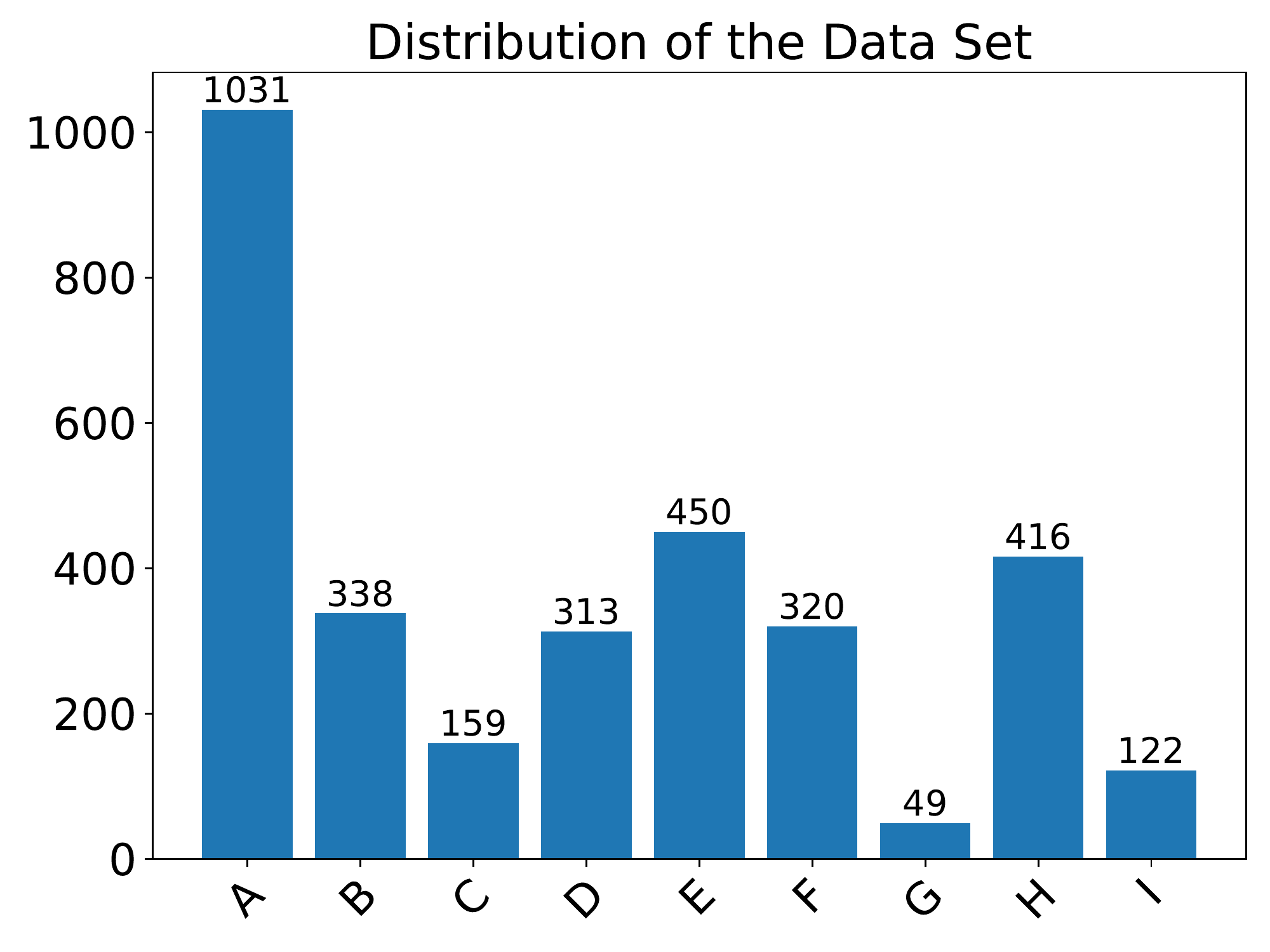}
    \caption{Histogram showing the distribution of the data set. Notice the large data imbalance, especially when comparing vehicle class `G' to vehicle class `A'.}
    \label{fig:data_dist}
\end{figure}

In this section, we provide further details of the data set. Figure \ref{fig:train_x} shows what the input domain looks like and Figure \ref{fig:data_dist} is a histogram showing the total data distribution. 

\acN{Include bar chart of data distribution to demonstrate imbalance and mention in the text that this is a difficult task?}

\subsection{Hyperparameter Optimisation}
The hyperparameters of all approaches were found via Bayesian optimisation (BO). For novel symmetric split HMC, we performed BO over vanilla HMC with a smaller subset of the data to reduce the computation time. 

\paragraph{Stochastic Gradient Descent: } Learning rate $=0.0103$; momentum $=0.9$; epochs $=209$; weight decay $=0.0401$; batch size $=512$.

\paragraph{SGLD: } Learning rate $=0.0182$; prior standard deviation $=0.7431$; epochs $=400$; batch size $=512$; burn $= 200$.

\paragraph{Stochastic Gradient HMC: } Learning rate $=0.0076$; prior standard deviation $=0.1086$; epochs $=1850$; friction term $=0.01$; batch size $=512$; burn $= 150$.

\paragraph{Novel Symmetric Split HMC: } $L=11$; $\epsilon = 4.96e^{-6}$; $\mathbf{M}=2e^{-5}\mathbf{I}$; $p(\bm{\omega}) = \mathcal{N}(\mathbf{0}, \tau^{-1} \mathbf{I})$, (with $ \tau = 100$); number of splits $=2$ (each of batch size 939); number of samples $= 3000$; burn $= 300$.

\subsection{Effect of the prior}

\acN{Talk about how small prior leads to the overconfidence collapse. Maybe show some uncertainty plots. Also state that the Log-Posterior Density is unnormalised so the absolute magnitude is not to be taken to mean too much but the trend lines are useful for indicating stability.}

\begin{figure}[h!]
    \centering
    \begin{subfigure}[b]{0.32\textwidth}
    \centering
        \includegraphics[width=\textwidth]{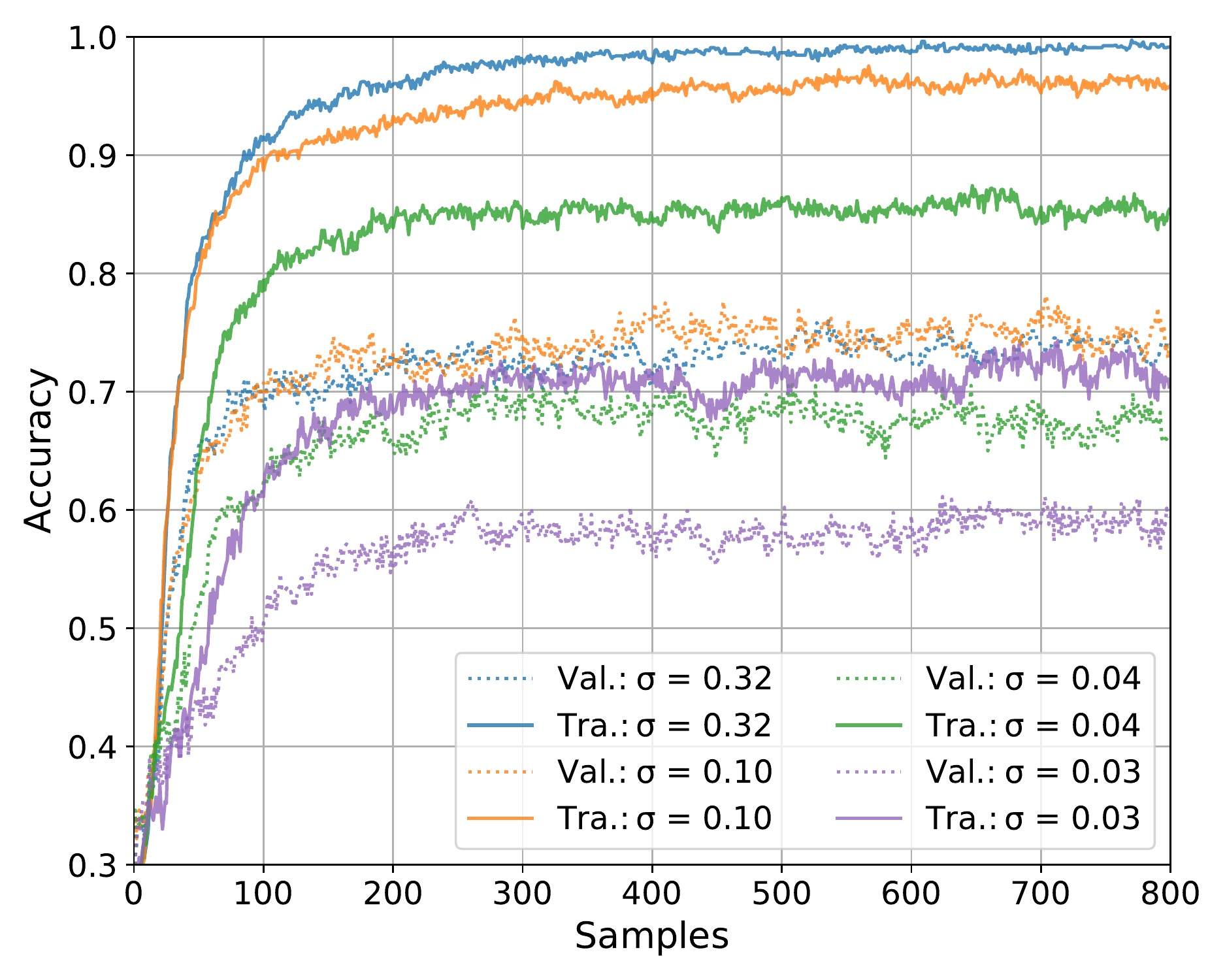}
        \caption{Accuracy}
        \label{fig:prior_comp_acc}
    \end{subfigure}
    \hfill 
    \begin{subfigure}[b]{0.32\textwidth}
    \centering
        \includegraphics[width=\textwidth]{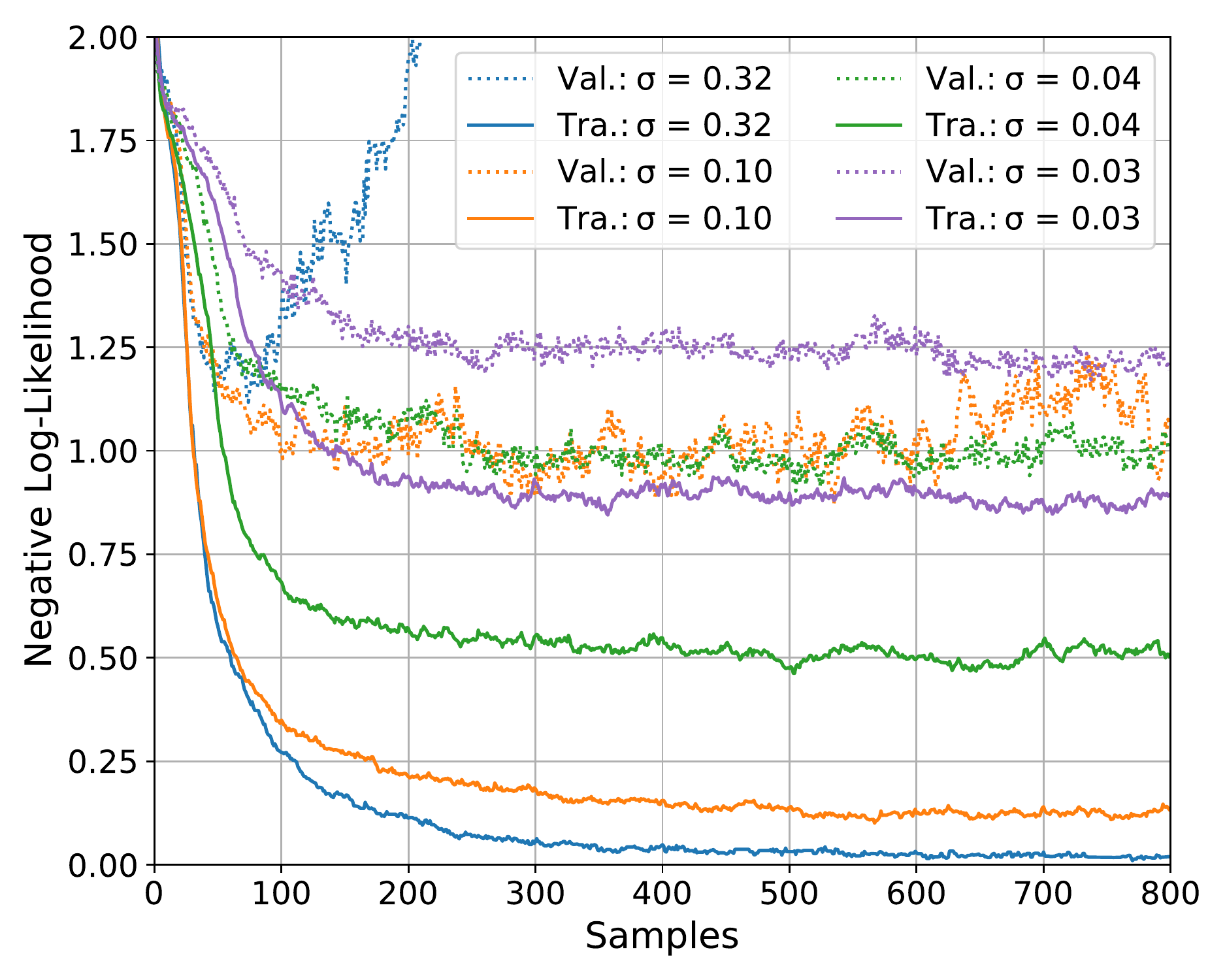}
        \caption{Mean Negative Log-Likelihood}
        \label{fig:prior_comp_nll}
    \end{subfigure}
    \hfill 
    \begin{subfigure}[b]{0.32\textwidth}
    \centering
        \includegraphics[width=\textwidth]{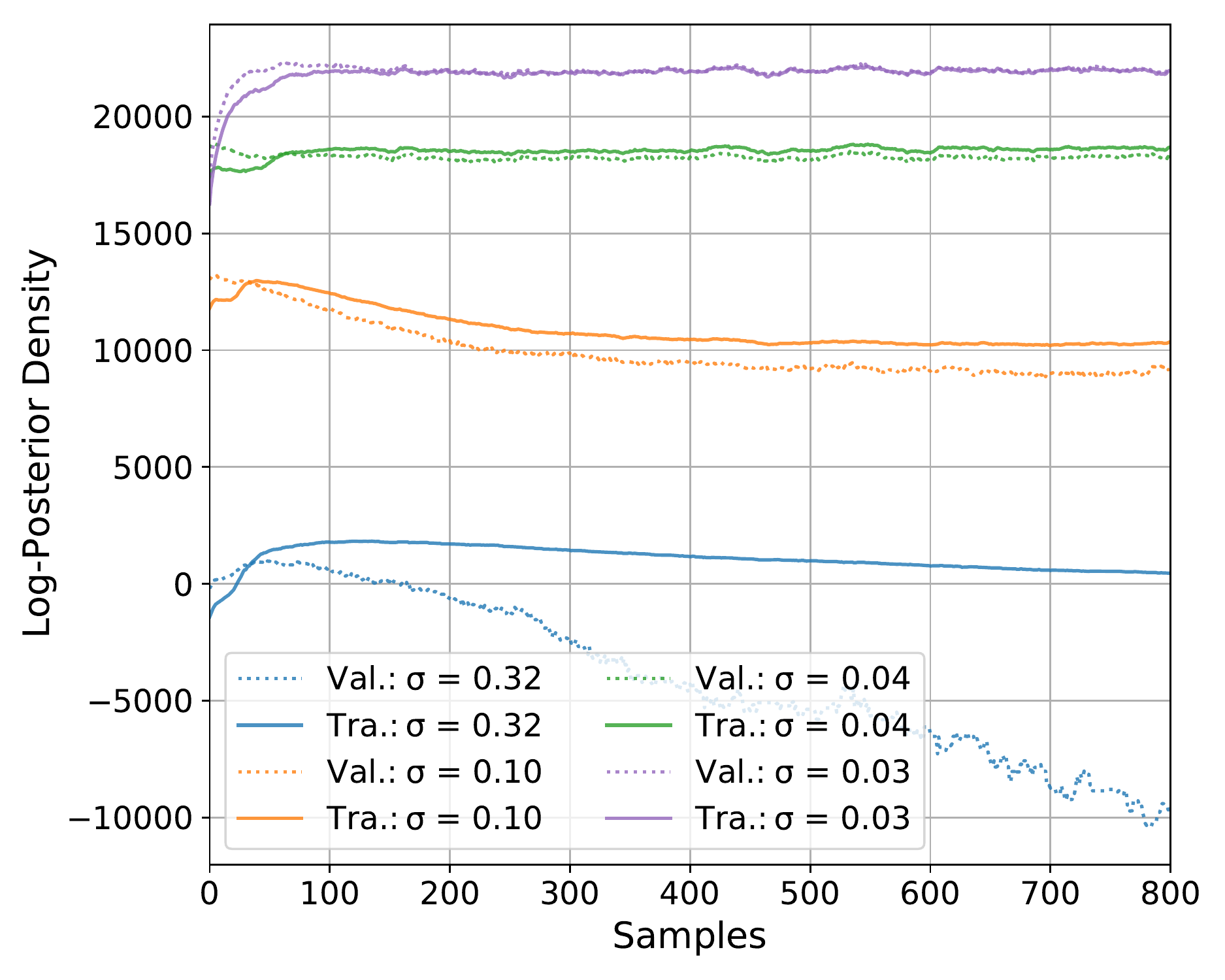}
        \caption{Log-Posterior Density}
        \label{fig:prior_comp_lpd}
    \end{subfigure}
    \caption{A performance comparison by varying the strength of the prior over the acoustic vehicle classification data set. We show the (a) Accuracy, (b) Negative Log-Likelihood (NLL), and (c) Log-Posterior Density. The curves are shown for $\sigma = 0.32,0.10,0.04,$ and $0.03$, where the prior $p(\bm{\omega}) = \mathcal{N}(\mathbf{0},\sigma^2 \mathbf{I}) $. When $\sigma$ is too large, the samples achieve almost $100 \%$ accuracy and zero NLL in the training data. However this is at the severe cost of the validation performance (see blue curves). When $\sigma$ is too small, the strength of the prior prevents the model from fitting to the data, with a lower accuracy and a higher NLL (see purple curves). The Log-Posterior Density acts as a good proxy for indicating convergence, by displaying the over-fitting collapse of $\sigma = 0.32$, where both the training and validation curves do not plateau.}\label{fig:prior_comparison}
\end{figure}

We use this section as an opportunity to demonstrate the effect of the prior on the classification results. Each weight in our network has a univariate Gaussian prior with a variance of $\sigma^2$ (i.e. $p(\bm{\omega}) = \mathcal{N}(\mathbf{0},\sigma^2 \mathbf{I}) $). We perform four experiments over the 
acoustic vehicle classification data, where $\sigma =0.32,0.10,0.04,$ and $0.03$ are used for each implementation.\footnote{Corresponding to precisions of $10, 100, 500,$ and $1000$ respectively.}
Figure \ref{fig:prior_comparison} shows the importance of carefully selecting the prior. Setting $\sigma$ to the larger (more flexible) value of $0.32$ leads to over-fitting.  For example in Figure \ref{fig:prior_comp_acc}, the solid blue curve yields near-perfect accuracy over the training data, with the validation curve also displaying a good accuracy performance. However this model is misspecified, which can easily be seen from the validation Negative Log-Likelihood (NLL) performance of Figure \ref{fig:prior_comp_nll}, which rapidly increases after approximately $100$ samples (see dotted blue curve). This misspecification is especially obvious, when we plot the Log-Posterior Density in Figure \ref{fig:prior_comp_lpd}. Unlike the accuracy and the NLL, the Log-Posterior Density indicates the model is performing poorly from simply observing the performance over training data, where we see the solid blue curve continuing to decrease with the number of samples (and not stabilising at a value like in the other settings).
These three indicators are especially important, as simply relying on accuracy would make it hard to distinguish between the best and worst performing models. 

\subsection{Classification Performance: Supplementary Plots}\label{ap:clas}

\begin{figure}[h]
\centering
    \includegraphics[width=0.8\columnwidth]{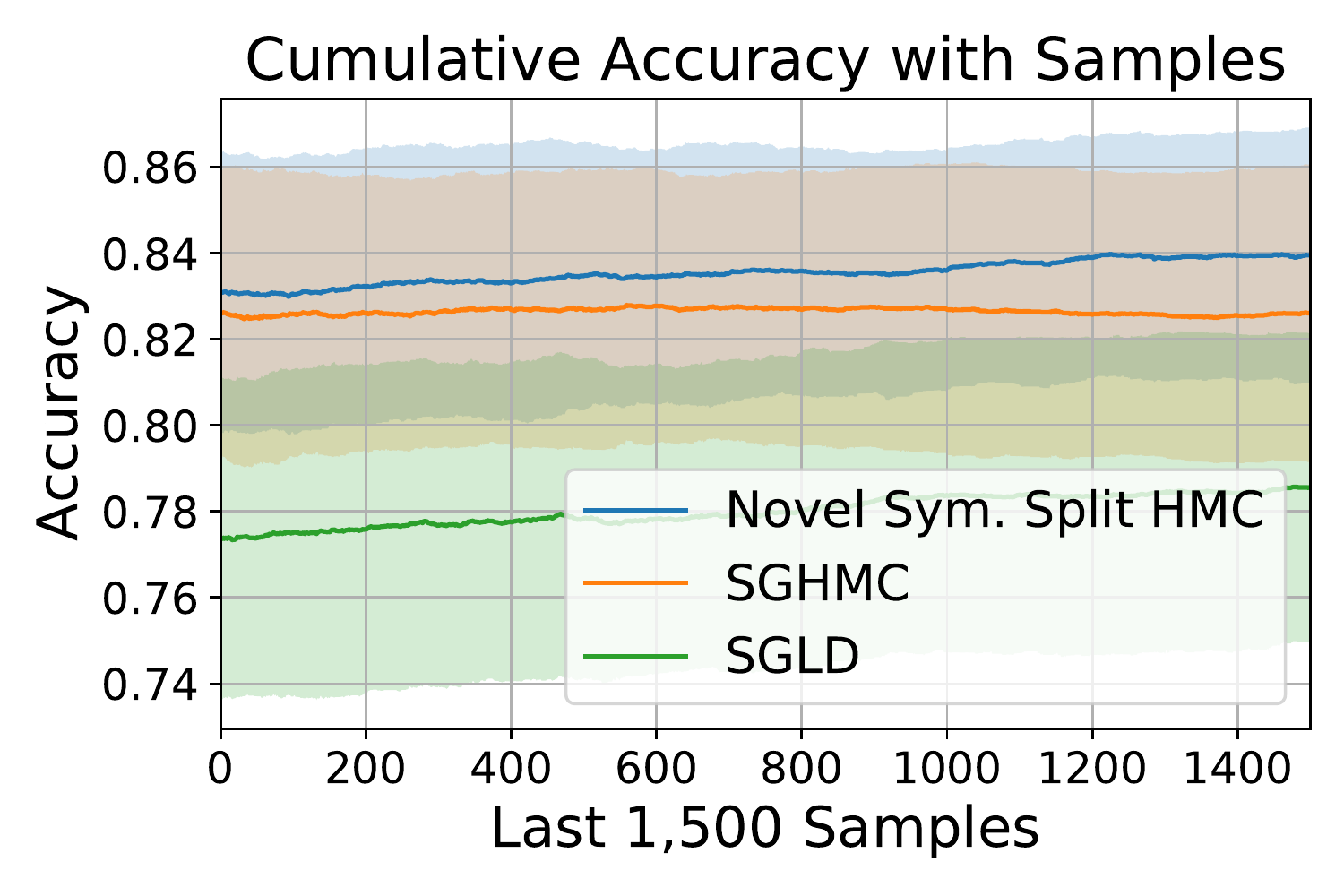}
    \caption{Cumulative accuracy of the ensemble of model samples. The standard deviation is over the cross-validation splits. The accuracy at each step is calculated by comparing the true label with $\max_c \mathbb{E}_{\bm{\omega}}[p(\mathbf{y}^*=c \vert \mathbf{x}^*)]$, where the expectation is over the samples up until that point. Novel symmetric split HMC continues to improve with the materialised number of samples.}
    \label{fig:cum_acc}
\end{figure}

We can also look at how the performance of the accuracy changes with the number of samples. Figure \ref{fig:cum_acc} shows the cumulative accuracy as the number of samples increases. In particular, we show the last $1{,}500$ samples for all MCMC schemes. Note that SGLD and SGHMC plateau at a lower mean accuracy.

\end{document}